\documentclass[10pt,twocolumn,letterpaper]{article}

\usepackage{cvpr}              
\definecolor{cvprblue}{rgb}{0.21,0.49,0.74}
\usepackage[pagebackref,breaklinks,colorlinks,allcolors=cvprblue]{hyperref}
\usepackage{url}            
\usepackage{booktabs}       
\usepackage{amsfonts}       
\usepackage{nicefrac}       
\usepackage{microtype}      
\usepackage{xcolor}         
\usepackage{makecell}
\usepackage{graphicx}

\usepackage{multirow}
\usepackage{pifont}
\usepackage{threeparttable}
\usepackage{amsmath}
\usepackage{enumitem}
\usepackage[accsupp]{axessibility}

\def\paperID{7633} 
\def\confName{CVPR}
\def\confYear{2026}

\title{Ultra-Fast Neural Video Compression}

\author{Jiahao Li$^1$, Wenxuan Xie$^1$, Zhaoyang Jia$^2\thanks{Done during Zhaoyang Jia's internship at Microsoft Research Asia.}$, \: Bin Li$^1$, Zongyu Guo$^1$, Xiaoyi Zhang$^1$, Yan Lu$^1$\\
$^1$Microsoft Research Asia, $^2$ University of Science and Technology of China\\
{\tt\small \{li.jiahao,wenxie,libin,zongyuguo,xiaoyizhang,yanlu\}@microsoft.com,} {\tt\small jzy\_ustc@mail.ustc.edu.cn}
}

\begin{document}
\maketitle
\begin{abstract}
While neural video codecs (NVCs) have demonstrated superior compression ratio, their prohibitive computational complexity remains a critical barrier to real-world deployment. This paper introduces a chunk-based coding framework designed to significantly improve the rate-distortion-complexity trade-off. Instead of processing frames sequentially, our approach encodes a chunk of multiple frames into a single compact latent representation and decodes them simultaneously. This is enabled by cross-frame interaction modules for joint spatial-temporal modeling and frame-specific decoders for parallel reconstruction. This paradigm not only dramatically enhances coding throughput but also facilitates more effective modeling of long-term temporal correlations. To further boost speed, we propose a streamlined entropy coding mechanism that consolidates bit-stream interactions into a single step, substantially reducing decoding overhead. Building on these innovations, we present DCVC-UF (Ultra-Fast), a new NVC that sets a new SOTA in performance. Our experiments show that DCVC-UF can achieve ultra-fast encoding and decoding speeds, significantly outperforming previous leading codecs. DCVC-UF serves as a notable landmark in the journey of NVC evolution. The code is at \url{https://github.com/microsoft/DCVC}.
\end{abstract}    
\section{Introduction}
\label{sec:intro}

Neural video codecs (NVCs) have emerged as transformative technologies, offering unprecedented capabilities in removing video redundancy. 
Recent works \cite{DCVC-FM, qi2024long, DCVC-RT,chen2023neural, kwan2024nvrc, gao2025givic, ma2025diffusion,qi2025generative, xue2025single} have driven rapid progress in improving compression ratios, enabling NVCs to surpass conventional codecs such as H.266/VTM \cite{VTM}. Despite these advances, the practical deployment of NVCs still faces significant challenges due to their substantial complexity in encoding or decoding. 
Consequently, achieving a better trade-off among rate, distortion, and complexity is a critical research direction.

\begin{figure}[t]
		\begin{center}
			\includegraphics[width=1.07\linewidth]{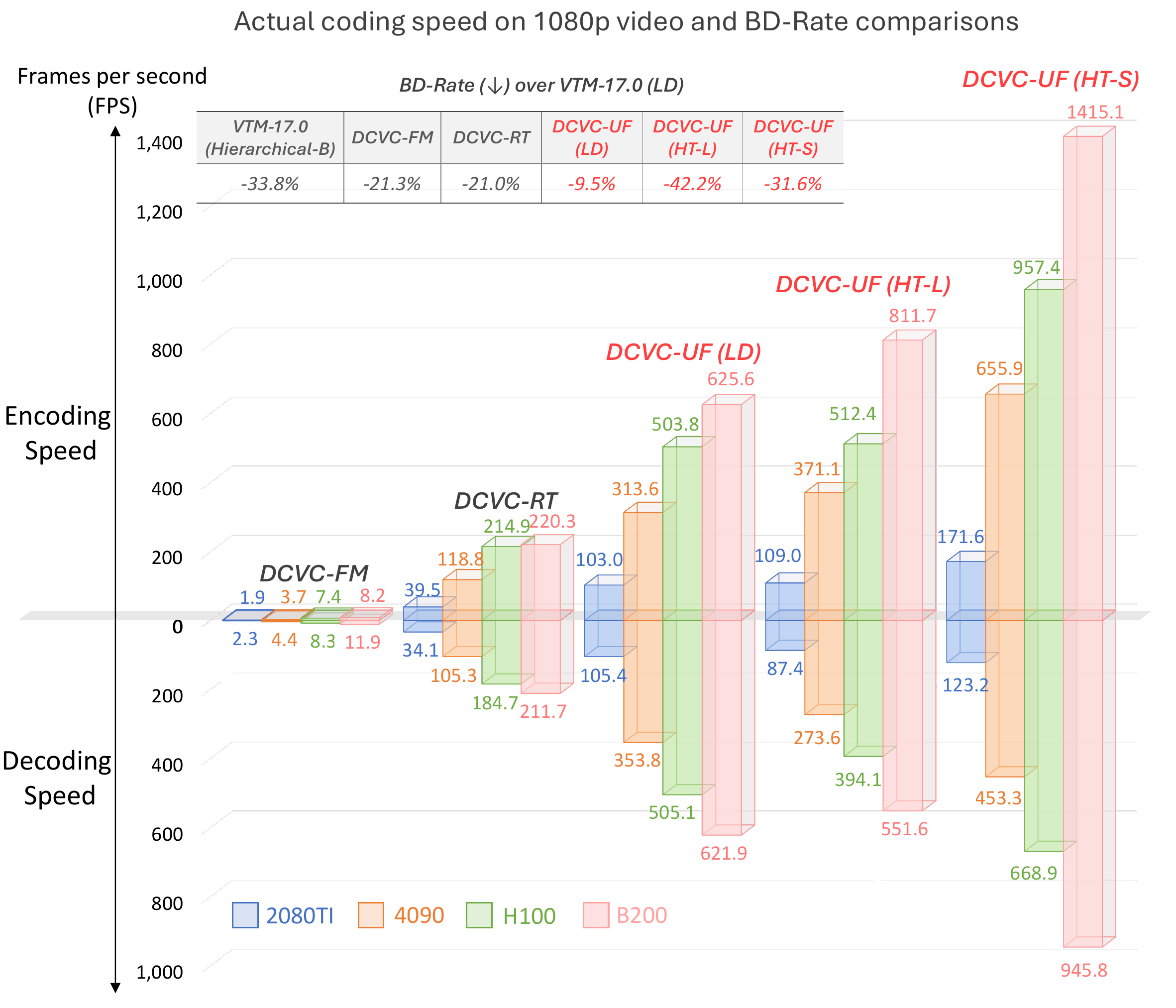}
		\end{center}
		\vspace{-0.5cm}
    \caption{Encoding and decoding speed with actual bit-stream writing and reading on $1920\times1080$ videos across different GPUs. Our DCVC-UF models achieve  unprecedented encoding and decoding speeds, demonstrating strong scalability and advanced rate-distortion-complexity trade-off on general-purpose GPUs.}
		\vspace{-6mm}
		\label{fig_first_performance}
\end{figure}
In response, recent approaches explore implicit neural representation (INR) \cite{sitzmann2020implicit} or 
 Gaussian Splatting \cite{kerbl20233d}, where each video is overfitted into implicit parameters \cite{chen2021nerv, kim2024c3} or explicit 2D Gaussians \cite{gupta2025neural,chen2025versatile}. Both paradigms offer low decoding complexity. However, as they need extensive online optimization for each video, their encoding complexity is quite high. For instance, \cite{gupta2025neural} reports that the encoding times for INRs are usually in the order of $10^{-3}$ FPS.
 
 Another direction to improve the compression ratio within limited computational budgets is to explore more efficient spatial-temporal correlation modeling. For traditional codec H.266/VTM, the hierarchical-B coding can achieve an average of 33.8\% bitrate saving over the low-delay coding by introducing the bidirectional temporal prediction in a GOP (group of pictures). So, recent NVCs \cite{yang2020learning, chen2023b,sheng2025bi,jiang2025biecvc} also follow a similar design to traditional hierarchical-B coding. However, they still operate on a frame-by-frame basis, where each frame relies on explicit motion vector for temporal prediction. The motion vector only captures the pixel displacement between only two frames, and cannot represent temporal correlation across multiple frames. Using different reference frames necessitates different motion vectors.
 In addition, motion vector struggles to handle the complex video dynamics or new contents \cite{gao2024implicit}, incurs non-trivial additional bitrate cost, and especially introduces the substantial system complexity. Hence, the capability of these NVCs' to balance rate, distortion, and complexity efficiently remains limited.

\begin{figure}[t]
		\begin{center}
			\includegraphics[width=1.07\linewidth]{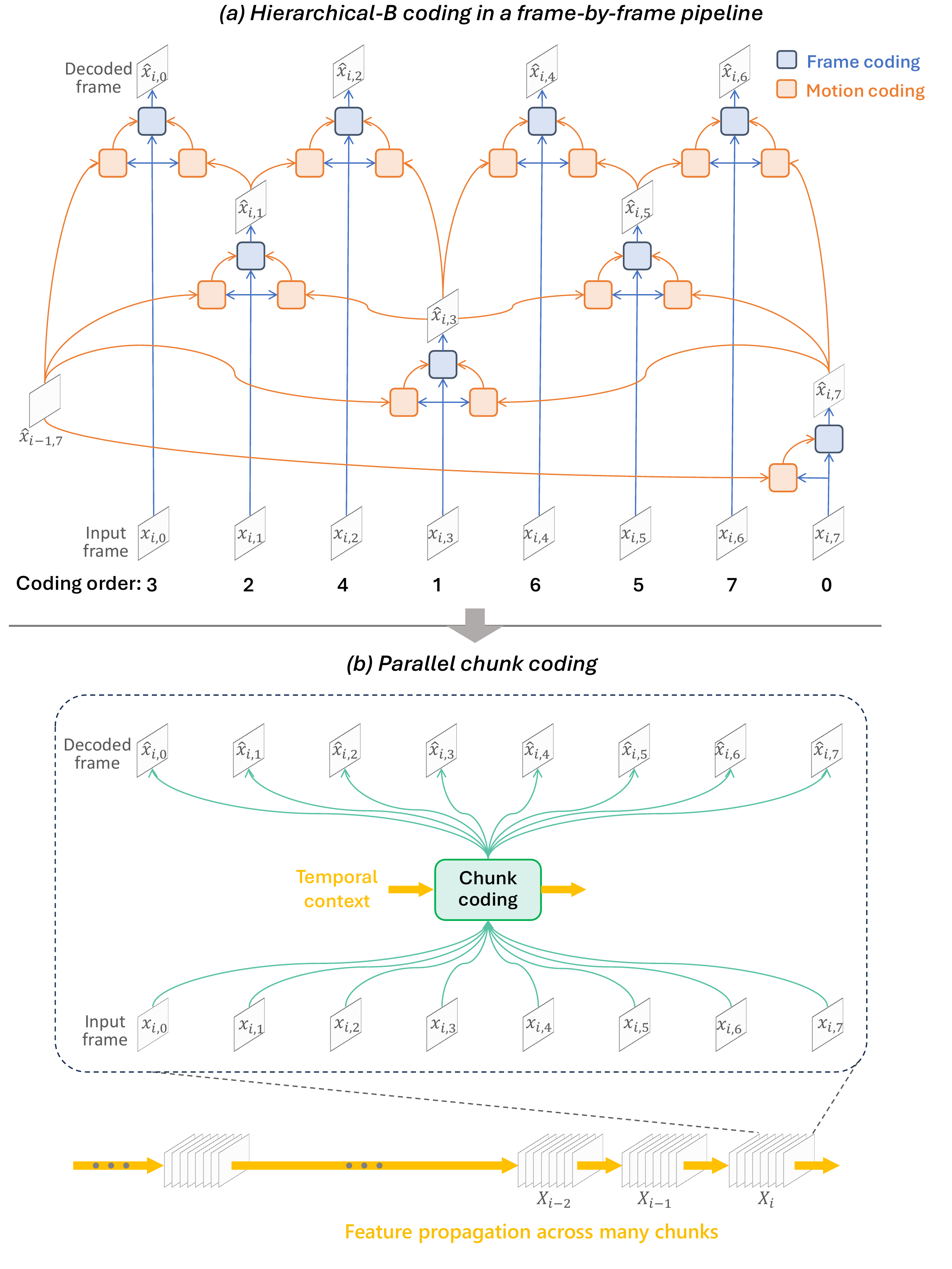}
		\end{center}
		\vspace{-0.6cm}
		\caption{Comparison of coding paradigms, illustrated with an 8-frame example. (a) The commonly-used hierarchical-B coding operates frame-by-frame, particularly relying on the complex motion coding. Here we show two reference frame example for B frames. If using more reference frames, more corresponding motion codings  are needed, increasing non-trivial cost. Moreover, the coding order adheres to a rigid, pre-defined hierarchy, and is not learnable. (b) Our chunk coding processes all frames of a chunk in parallel to automatically learn the spatial-temporal correlation. It eliminates the explicit motion, significantly enhances the throughput, and yet enables more efficient long-term temporal modeling.}
		\vspace{-5mm}
		\label{fig_framework_change}
\end{figure}

In this paper, motivated by the spatial-temporal autoencoder \cite{habibian2019video} and the motion vector-free DCVC-RT \cite{DCVC-RT}, we propose a chunk-based coding framework to address the aforementioned challenges. As shown in Fig. \ref{fig_framework_change}, instead of sequentially processing the video frame-by-frame, our approach divides the video into non-overlapping chunks of multiple frames. 
All frames within each chunk are then encoded into compact latent representations and decoded back simultaneously, which is designed to maximize coding throughput. 
Within this chunk-based framework, our architecture employs cross-frame interaction modules to jointly and implicitly model spatial-temporal correlations across all frames. Complementing this, a set of frame-specific decoders works in parallel to reconstruct each frame, adaptively tailoring the synthesis process to individual frame characteristics.
This paradigm facilitates a more holistic compression strategy and also can maximize coding throughput.
It amplifies the advancements of the motion vector-free design from DCVC-RT, removing the costly, iterative, and sequential motion estimation, motion entropy coding, and compensation processes between many frame pairs. Our approach significantly reduces the operational complexity, such as memory I/O and function call overhead, which are critical bottlenecks for practical coding speed.

Our chunk-based coding also enables more efficient modeling of long-term temporal context. One important reason why the previous SOTA DCVC series \cite{DCVC-DC,DCVC-FM,DCVC-RT} surpasses leading traditional codecs is the enabling of the feature propagation mechanism in the latent space, which implicitly captures temporal correlations across multiple frames through joint training. Notably, DCVC-FM \cite{DCVC-FM} shows that the compression ratio can be significantly boosted by increasing the training video length from 7 to 32 frames. However, extending the training to longer video sequences is challenging because each frame's separate latent representation incurs substantial training costs. In contrast, our framework encodes all frames in a chunk into a single compact latent, significantly reducing the latent size for a video. It allows for training on much longer video sequences within limited computational budgets, thereby facilitating the exploration of long-term temporal correlations to improve the chunk latent generation and the corresponding distribution estimation.

To further accelerate practical coding speed during the conversion between chunk latents and bit-streams, we introduce a streamlined entropy coding mechanism. Early NVCs \cite{DCVC, lu2020end} typically employ auto-regressive decoding \cite{minnen2018joint}, which is inherently slow due to its sequential nature. The recent quadtree partition-based method in \cite{DCVC-DC} reduces decoding steps to four by leveraging decoded partition latents to  estimate the means and scales for the next partition in parallel. However, even this four-step interaction with the bit-stream incurs notable operational overhead, especially in real-time scenarios. Our proposed method further simplifies this process by estimating the scales for all partitions in a single step, while retaining the four-step estimation for the means to keep the spatial-channel correlation modeling capability. Since bit-stream decoding depends only on the scales, this allows us to consolidate bit-stream interactions into a single step, substantially reducing operational cost and improving bit-stream decoding efficiency.

Together, these advancements culminate in our proposed NVC, named DCVC-UF (Ultra-Fast), which builds upon the DCVC series to deliver exceptional encoding and decoding speeds. To accommodate diverse application scenarios, we introduce several configurations of DCVC-UF. Fig. \ref{fig_first_performance} shows the performance comparison, where the VTM (Low-Delay, LD) is as the anchor for BD-Rate calculation.
When the chunk has multiple frames, we can achieve High-Throughput (HT) coding yet with high compression ratio. Our large version DCVC-UF (HT-L) saves an average of 42.2\% bitrate, with achieving 371.1 encoding and 273.6 decoding FPS for 1080p video with 4090 GPU. Our small version DCVC-UF (HT-S) achieves 31.6\% bitrate saving. The encoding and decoding speeds are boosted to  655.9 and 453.3, respectively. 
These two models will introduce the delay related to the chunk size, analogous to the hierarchical-B coding manner. So, to meet the low-delay requirement, we can also set the chunk size to 1 frame, i.e., DCVC-UF (LD),  which can achieve 9.5\% bitrate saving, yet achieve 313.6 encoding and 353.8 decoding FPS with 4090 GPU. 
Unlike traditional codecs requiring bespoke hardware, our NVC framework is built on general-purpose GPUs, allowing it to automatically benefit from rapid advancements in AI accelerators without re-engineering. This inherent scalability is demonstrated as the speeds of our models  consistently improve across GPU generations, from consumer cards to datacenter accelerators like the B200, where DCVC-UF (HT-S) sets a new throughput record of 1415.1 encoding and 945.8 decoding FPS for 1080p. As consumer GPUs advance in the future, their speeds will also rise automatically.

Our main contributions are summarized as follows:
\begin{itemize}
    \item We propose a chunk-based coding framework. It encodes a chunk of frames into a single compact latent and decodes back simultaneously, leveraging cross-frame interaction and frame-specific decoders. This design significantly enhances coding throughput and enables more efficient modeling of long-term temporal context.
    \item We design a streamlined entropy coding mechanism that decouples the estimation of scales and means for latent partitions. This allows bit-stream interactions to be consolidated into a single step, substantially reducing operational overhead and accelerating practical decoding speed.
    \item Extensive experiments demonstrate that our model, DCVC-UF, establishes a new record of the rate-distortion-complexity performance, significantly outperforming previous SOTA codecs across various settings.
\end{itemize}

\section{Related Work}
\subsection{Low-Delay Neural Video Compression}
Low-delay coding constrains the coding of the current frame to reference only previously decoded frames in temporal order, suitable for real-time communication applications. Many methods \cite{lu2019dvc, hu2020improving,Rippel_2021_ICCV,lu2020end,lin2020m,agustsson2020scale, liu2020neural, liu2023mmvc, ma2024uncertainty} adopt a residual coding inspired by traditional codecs, which requires a complex motion estimation, entropy coding, and compensation pipeline. The emerging conditional coding \cite{liu2020conditional, ladune2021conditional,canfvc,vct, DCVC,sheng2021temporal, DCVC-HEM, qi2023motion, DCVC-DC, DCVC-FM, qi2024long} shows larger potential as the temporal context is not limited to pixel-domain prediction but can be any flexible feature. However, most of them still suffer from limited coding speed, as they often incorporate complex modules—particularly those related to motion processing. While some works \cite{van2024mobilenvc, le2022mobilecodec, tian2023towards} prioritize acceleration, their compression performance lags behind leading approaches.

\subsection{Delay-Relaxed Neural Video Compression}
Relaxing the frame reference constraint enables a larger design space, but increases delay. It is suited for delay-insensitive scenarios like offline storage and video streaming.

\textbf{Hierarchical-B Coding.} Drawing inspiration from the significant compression gains of hierarchical-B coding over low-delay configurations in traditional codecs, several recent NVCs \cite{Djelouah_2019_ICCV,yang2020learning, chen2023b,sheng2025bi,jiang2025biecvc} have adopted analogous coding structures. This allows frames to reference both past and future frames for improved prediction. However, these methods still operate on a frame-by-frame basis, relying on explicit motion vectors to align two frames. To mitigate the bitrate overhead of motion vectors, some approaches \cite{chen2023b,sheng2025bi} introduce complex motion vector prediction modules, which in turn increase both computational and operational costs.

\textbf{Online Optimization-based Coding.} This paradigm trains a specialized model for each video instance. INR-based methods \cite{chen2021nerv, kwan2023hinerv, chen2023hnerv, kwan2024nvrc, kim2024c3, gao2025pnvc} overfit a small neural network to represent a video, and  decode frames by querying the network with coordinates. However, INRs are inefficient for high-resolution video \cite{gupta2025neural}. Consequently, recent works \cite{liu2023exploration, wang2025gsvc, gupta2025neural, lee2025gaussianvideo, chen2025versatile} explore explicit representations using Gaussian Splatting \cite{kerbl20233d}. This method associates Gaussian parameters with video regions, enabling scalable representation and faster rendering \cite{gupta2025neural}. While both approaches achieve high decoding speed, their per-video online optimization results in extremely high encoding complexity.

\begin{figure*}[t]
  \centering
    \includegraphics[width=\linewidth]{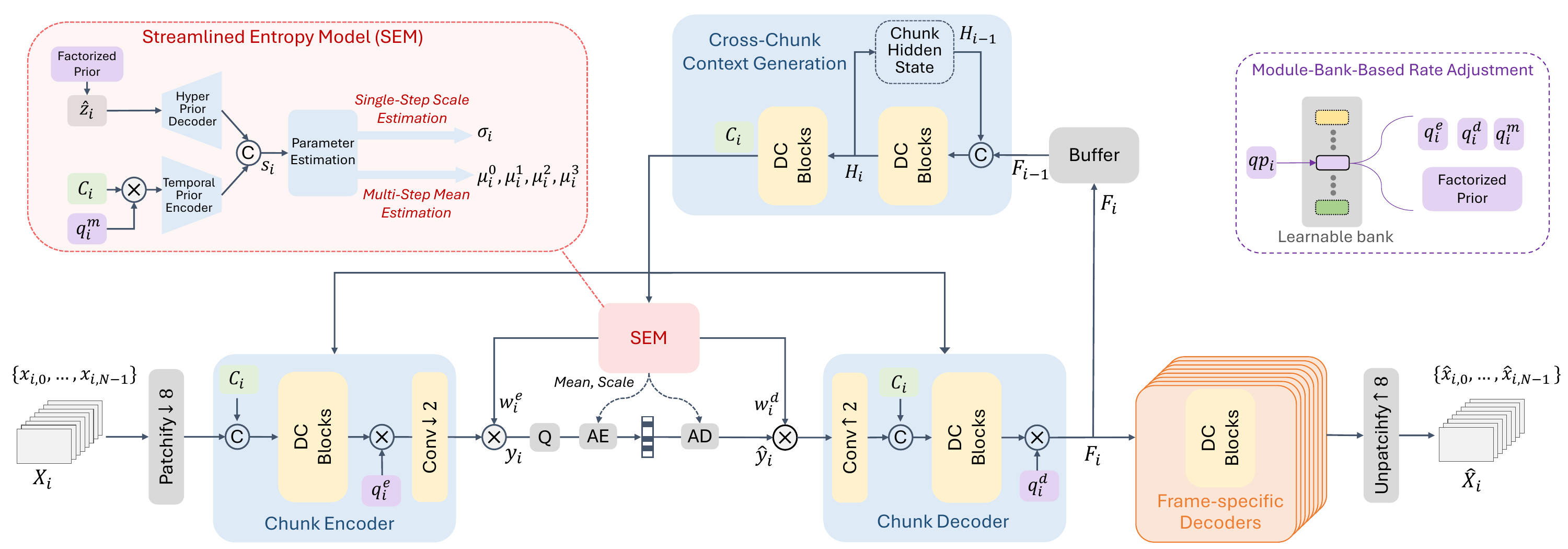}
    \vspace{-4mm}
    \caption{Framework overview of our DCVC-UF. DC Block, Q, AE and AD represent depth-wise convolution block, quantization, arithmetic encoder and decoder, respectively. After the patchify, the input chunk $X_{i}$ (comprising $N$ frames) directly encoded and decoded into feature $F_{i}$, conditioned on the temporal chunk context $C_{i}$. $F_{i}$ is then reconstructed into pixel domain using the frame-specific decoders. $qp_{i}$ is the input quantization parameter. The number of DC Blocks for each module is detailed in the supplementary material. In DCVC-UF, all frames in the chunk are processed in parallel to enable the high-throughput coding.
    }
    \vspace{-4mm}
  \label{fig_main_framwork}
\end{figure*}

\textbf{Spatial-Temporal Autoencoders.} In video generation, spatial-temporal autoencoders serve as powerful tokenizers, compressing raw pixels into a compact latent space to mitigate the prohibitive computational costs of generation in the pixel domain \cite{habibian2019video}. Recent works \cite{zheng2024open, zhao2024cv, chen2024od, wang2024omnitokenizer,yang2024cogvideox, kong2024hunyuanvideo, wu2025improved, wan2025wan} commonly employ configurations with spatial (e.g., 8x) and temporal (e.g., 4x) compression \cite{chen2025dc}. While primarily designed for generation, the underlying principle of converting raw video chunk into compact representations makes these autoencoders a promising foundation for developing efficient video codecs. Actually, early works \cite{habibian2019video,pessoa2020end} explored NVCs based on spatial-temporal autoencoder. However, \cite{habibian2019video,pessoa2020end} use a vanilla autoencoder to mainly learn the  the inner correlation within a single chunk. The correlation across chunks is ignored, leading to their limited compression ratio. 

Our work advances the spatial-temporal autoencoder paradigm for NVC. Within a chunk, unlike \cite{habibian2019video,pessoa2020end}, our autoencoder not only has cross-frame interaction modules for joint spatial-temporal modeling but also has frame-specific decoders tailoring the synthesis process to individual frame characteristics. Across different chunks, we build the efficient conditional coding, where temporal propagation is enabled to capture the implicit long-term correlation therein. In addition, we propose a streamlined entropy coding mechanism that consolidates bit-stream interactions into a single step, substantially accelerating the decoding. These make our NVC achieve significant rate-distortion-complexity trade-off advantage over \cite{habibian2019video,pessoa2020end} and other previous SOTA codecs.

\section{Proposed Method}
\subsection{Overview}

As depicted in Fig. \ref{fig_main_framwork}, our DCVC-UF is architected around a chunk-coding paradigm, building upon the DCVC series \cite{DCVC-DC, DCVC-FM, DCVC-RT}. The input video is first segmented into non-overlapping chunks. For a given chunk $X_i = \{x_{i,0}, \dots, x_{i,N-1}\}$ containing $N$ frames, the process begins by transforming it to 1/8 resolution via patch embedding. It is then conditioned on the temporal chunk context $C_i$, and fed into a chunk encoder. The role of the encoder is to distill the spatial-temporal information of the entire chunk into a compact latent representation $y_i$ efficiently. This latent is then quantized ($\hat{y}_i$) and efficiently converted into a bit-stream. During decoding, the process is reversed. The latent representation $\hat{y}_i$ is parsed from the bit-stream and fed to the chunk decoder, which generates a rich feature $F_i$. 
This feature serves a dual purpose: it is used by a set of parallel, frame-specific decoders to reconstruct the individual frames $\{\hat{x}_{i,0}, \dots, \hat{x}_{i,N-1}\}$, and it is also propagated to the next chunk to form the next temporal context. 
DCVC-UF originates from the low-delay NVC DCVC-RT \cite{DCVC-RT} which eliminates the explicit motion-related operations. DCVC-UF amplifies its advantage and enables high-throughput coding via our chunk-coding. DCVC-UF can boost the compression ratio of NVC to a new level with our frame-specific decoders (Sec. \ref{sec_decoder}) and efficient long-term correlation learning (Sec. \ref{sec_long}). In particular, our DCVC-UF also achieves unprecedented coding speed with our streamlined entropy model (Sec. \ref{sec_entropy_model}).

\subsection{Frame-Specific Decoders}
\label{sec_decoder}
Existing spatial-temporal autoencoders typically employ a unified decoder that applies identical reconstruction processes to all frames within a chunk. While this unified approach is straightforward to implement, it faces limitations when dealing with diverse contents. A single decoder must learn to handle all possible variations across different temporal positions, leading to a challenging optimization problem where the decoder needs to be a ``jack of all trades". This often results in suboptimal reconstruction quality, as the decoder cannot fully specialize for the distinct characteristics that may appear at different temporal positions within a chunk. To address these limitations, we additionally design frame-specific decoders in our chunk-based framework, where each frame index within the chunk is assigned its own dedicated decoder. As illustrated in Fig. \ref{fig_main_framwork}, after the chunk decoder generates the rich feature representation $F_i$ containing spatial-temporal information for all $N$ frames, we deploy $N$ distinct decoders operating in parallel. Each decoder specializes in reconstructing the frame at its corresponding position.

This design shares some similarities with Mixture of Experts (MoE) \cite{riquelme2021scaling} architecture, where specialized components handle different aspects of the content. In our case, each frame-specific decoder acts as an ``expert" for its temporal position, allowing the model to better adapt to varying video content characteristics. By distributing the reconstruction task across multiple specialized decoders, we can achieve several advantages: (1) Each decoder can focus on learning patterns most relevant to its position, reducing the complexity of individual decoder optimization; (2) The parallel architecture naturally aligns with our chunk-based processing, enabling simultaneous reconstruction without sequential dependencies; (3) The specialization allows for more efficient parameter utilization, as each decoder can allocate its capacity to the specific challenges of its assigned position rather than attempting to generalize across all positions.

\subsection{Efficient Long-Term Correlation Learning}
\label{sec_long}

One of the key advantages of our chunk-based coding framework is its ability to efficiently model long-term temporal correlations. Previous DCVC series have demonstrated that feature propagation mechanisms in the latent space can implicitly capture temporal correlations across multiple frames through joint training, which is a primary factor in their superiority over traditional codecs. Notably, DCVC-FM \cite{DCVC-FM} showed that compression ratio can be significantly improved by extending training video length from 7 to 32 frames, enabling the model to learn more comprehensive temporal dependencies. However, scaling to even longer sequence in frame-based approaches faces fundamental limitations: each frame requires its own latent representation, leading to large training cost. This constraint severely restricts the temporal context that can be practically leveraged during training.

Our chunk-based architecture fundamentally addresses this limitation by encoding all frames within a chunk into a single compact latent representation. This design dramatically reduces the total latent size for a video sequence. This compact representation enables training on much longer sequence (if the batch size is 1, it can be up to 1,024 frames at $512\times512$ spatial size, within 24GB GPU memory cost), letting the model capture long-term temporal correlation. The extended temporal context benefits both the chunk latent generation process and the entropy model's distribution estimation. During training, the model learns to identify and exploit recurring patterns, scene structures, and dynamics that span across multiple chunks. The propagated chunk context $C_i$ carries forward essential information, helping subsequent chunks achieve higher compression efficiency.

\subsection{Streamlined Entropy Model}
\label{sec_entropy_model}

\begin{figure}[t]
		\begin{center}
			\includegraphics[width=1.03\linewidth]{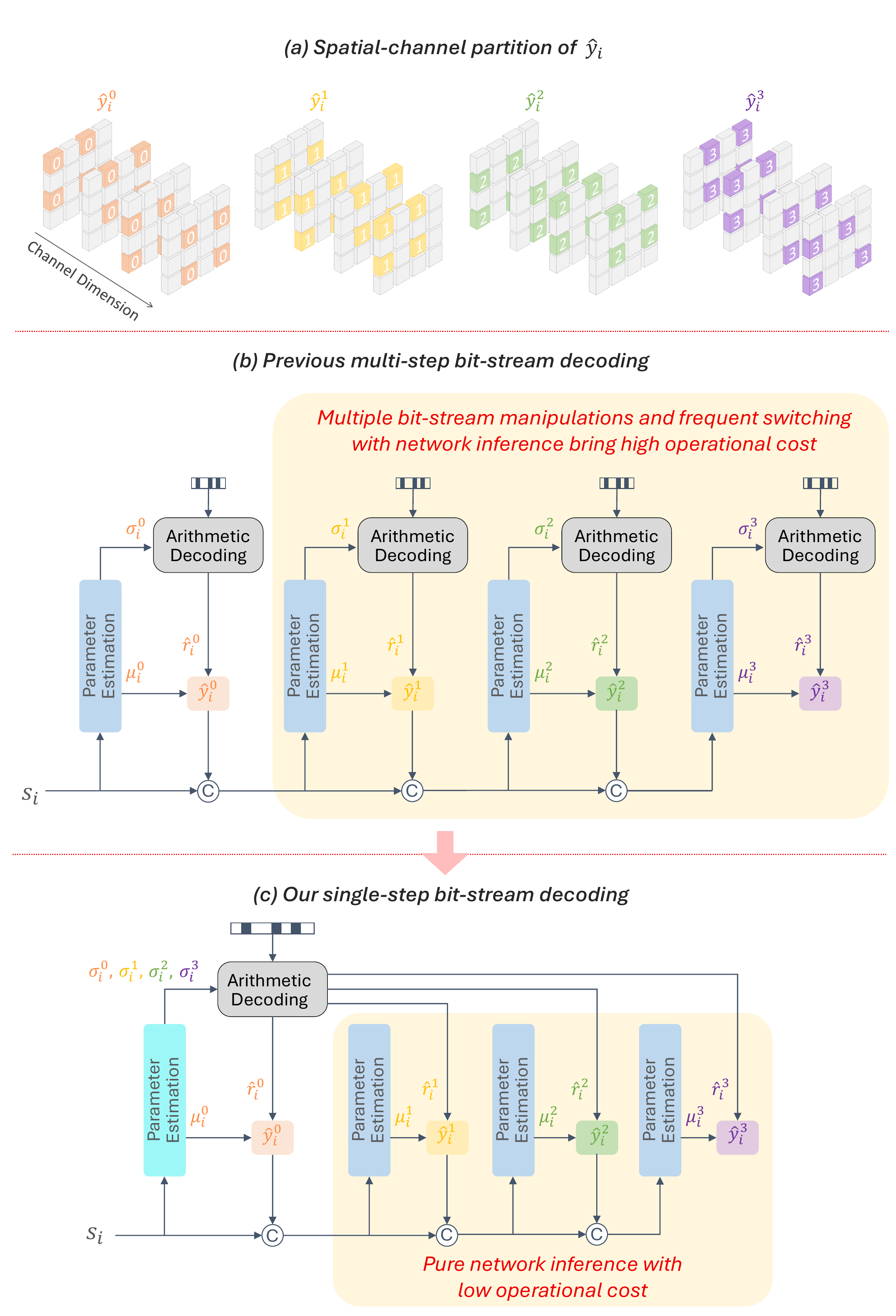}
		\end{center}
		\vspace{-0.4cm}
    \caption{ (a) A quadtree-like partition \cite{DCVC-DC} for $\hat{y}_{i}$ is adopted. (b) Previous methods require interleaved entropy decoding and parameter estimation, which hinders practical decoding speed. (c) Our streamlined entropy model consolidates bit-stream manipulations into a single step, substantially accelerating decoding.}
		\vspace{-4mm}
		\label{fig_entropy_model}
\end{figure}

\begin{table*}[t]
    \caption{BD-Rate (\%) comparison in YUV420 colorspace. All frames are tested. }
    \vspace{-0mm}
    \centering
    \renewcommand\arraystretch{1.8}
    \setlength{\tabcolsep}{3.5pt}
    \resizebox{0.9\linewidth}{!}{
    \begin{threeparttable}
    	\begin{tabular}{ l c c c c c c c c c }
            \toprule
    		\multirow{2}{*}{Method} & \multirow{2}{*}{UVG} & \multirow{2}{*}{MCL-JCV} & \multirow{2}{*}{HEVC B} & \multirow{2}{*}{HEVC C} & \multirow{2}{*}{HEVC D} & \multirow{2}{*}{HEVC E} & \multirow{2}{*}{Average} & \multicolumn{2}{c}{Coding Speed} \\
            ~ & ~ & ~ & ~ & ~ & ~ & ~ & ~ & Enc. & Dec. \\
            \hline    
            \textit{Low-Delay (LD) Codecs}\\
            \hline
            VTM-17.0 (LD)                       & $0.0$   & $0.0$   & $0.0$   & $0.0$   & $0.0$   & $0.0$   & $0.0$  & $0.01$ FPS  & $23.6$ FPS\\
            \hline
            HM-16.25 (LD)                       & $40.1$  & $48.6$  & $47.6$  & $41.0$  & $34.5$  & $42.8$  & $42.4$  & $0.05$ FPS & $39.6$ FPS\\
            \hline
            DCVC-DC                             & $6.5$   & $-4.4$  & $13.1$  & $-3.4$  & $-14.8$ & $90.2$  & $14.5$  & $2.3$ FPS  & $2.9$ FPS \\
            \hline    
            DCVC-FM                             & $-16.8$ & $-8.0$  & $-15.4$ & $-30.2$ & $-37.5$ & $-20.2$ & $-21.3$ & $3.7$ FPS  & $4.4$ FPS  \\
            \hline
            DCVC-RT                             & $-24.0$ & $-14.8$ & $-16.6$ & $-21.0$ & $-27.3$ & $-22.4$ & $-21.0$ & $118.8$ FPS& $105.3$ FPS \\
            \hline    
            \textbf{DCVC-UF (LD)}               & $-15.3$ & $-0.3$  & $-3.3$ & $-6.5$ & $-16.6$ & $-15.0$   & $-9.5$ & $\mathbf{313.6}$ FPS & $\mathbf{353.8}$ FPS \\
            \hline    
            \textit{Delay-Relaxed Codecs}\\
            \hline
            HM-16.25 (Hierarchical-B)     & $4.9$   & $17.3$  & $12.6$  & $11.3$  & $3.2$   & $1.6$   & $8.5$   & $0.06$ FPS & $40.0$ FPS \\  
            \hline
            VTM-17.0 (Hierarchical-B)     & $-34.0$ & $-30.4$ & $-35.4$ & $-32.4$ & $-32.5$ & $-38.1$ & $-33.8$ & $0.01$ FPS & $23.1$ FPS \\  
            \hline
            \textbf{DCVC-UF (HT-S)}      & $-28.8$ & $-12.9$ & $-17.6$ & $-29.4$ & $-42.2$ & $-58.8$ & $-31.6$ & $\mathbf{655.9}$ FPS & $\mathbf{453.3}$ FPS \\        
            \hline
           \textbf{DCVC-UF (HT-L)}       & $-39.6$ & $-24.4$ & $-33.3$ & $-41.2$ & $-51.7$ & $-63.2$ & $-42.2$ & $\mathbf{371.1}$ FPS & $\mathbf{273.6}$ FPS \\
            \bottomrule
    	\end{tabular}
        \begin{tablenotes}
            \item  Intra-period=--1 for all codecs and settings. The coding speeds of NVCs are tested on $1920 \times 1080$ videos with 4090 GPU. 
        \end{tablenotes}
    \end{threeparttable}
	}
  \label{tab:compare_yuv_allf}
  \vspace{-0mm}
\end{table*}

Efficient entropy coding is crucial for achieving high practical coding speed, as it directly determines how quickly latent representations can be converted to and from bit-streams. Recently, the quadtree partition-based entropy coding \cite{DCVC-DC, DCVC-FM} was proposed to explore spatial-channel correlations efficiently, demonstrating significantly higher decoding speed than the famous auto-regressive model \cite{minnen2018joint}. As illustrated in Fig. \ref{fig_entropy_model} (a) and (b), the quantized latent $\hat{y}_{i}$ is divided into four partitions, where each partition's decoding depends on previously decoded partitions to estimate its distribution parameters (mean $\mu$ and scale $\sigma$).  However, even this four-step process still incurs substantial operational overhead. As highlighted in Fig. \ref{fig_entropy_model} (b), the repeated bit-stream manipulations involve multiple arithmetic decoding calls, memory I/O operations, and costly synchronization between arithmetic decoding operations and neural network inference. If the arithmetic decoding is performed on CPU, the cross-device switching and synchronization between CPU and GPU further exacerbate the operational burden.

To address these bottlenecks, we first revisit the entropy coding process. During encoding, the entropy model estimates the mean $\mu_i$ and scale $\sigma_i$ for the latent $y_i$, typically assuming a Gaussian distribution. After quantization via $\hat{r}_{i} = \text{round}(y_{i}-\mu_{i})$, the result $\hat{r}_{i}$ is arithmetically encoded using scale $\sigma_i$. During decoding, $\hat{r}_{i}$ is recovered using only $\sigma_i$, and the final latent is reconstructed as $\hat{y}_{i}=\hat{r}_{i}+ \mu_{i}$. The key insight is that bit-stream operations depend exclusively on scale parameters, which define the distribution width, while mean parameters merely shift the distribution center and can be applied post-decoding. This observation motivates us to decouple the estimation of means and scales, departing from the coupled approach in \cite{DCVC-DC, DCVC-FM}. As shown in Fig. \ref{fig_entropy_model} (c), our parameter estimation network takes the prior input $s_i$ (derived from hyper-prior $\hat{z}_{i}$ and temporal context $C_{i}$) and simultaneously predicts the mean $\mu_{i}^{0}$ for the first partition and the scales $\sigma_i$ for all four partitions in a single forward pass.

This architectural innovation enables a dramatic acceleration of the decoding pipeline. By eliminating sequential dependencies in scale estimation, we can perform arithmetic decoding for all partitions in one consolidated step, drastically reducing bit-stream manipulation overhead. We retain the four-step progressive estimation for means to preserve spatial-channel correlation modeling capacity, but since mean estimation requires no bit-stream interaction, it executes entirely on GPU without costly synchronization. This design minimizes memory transfer between processing units, eliminates multi-step decoding latency, and removes switching overhead between arithmetic operations and neural network inference.  Our streamlined entropy model allows for better GPU utilization, combined with our chunk-based coding framework, enables DCVC-UF to achieve unprecedented decoding speed.

\begin{figure*}[t]
\vspace{-4mm}
  \centering
    \includegraphics[width=\linewidth]{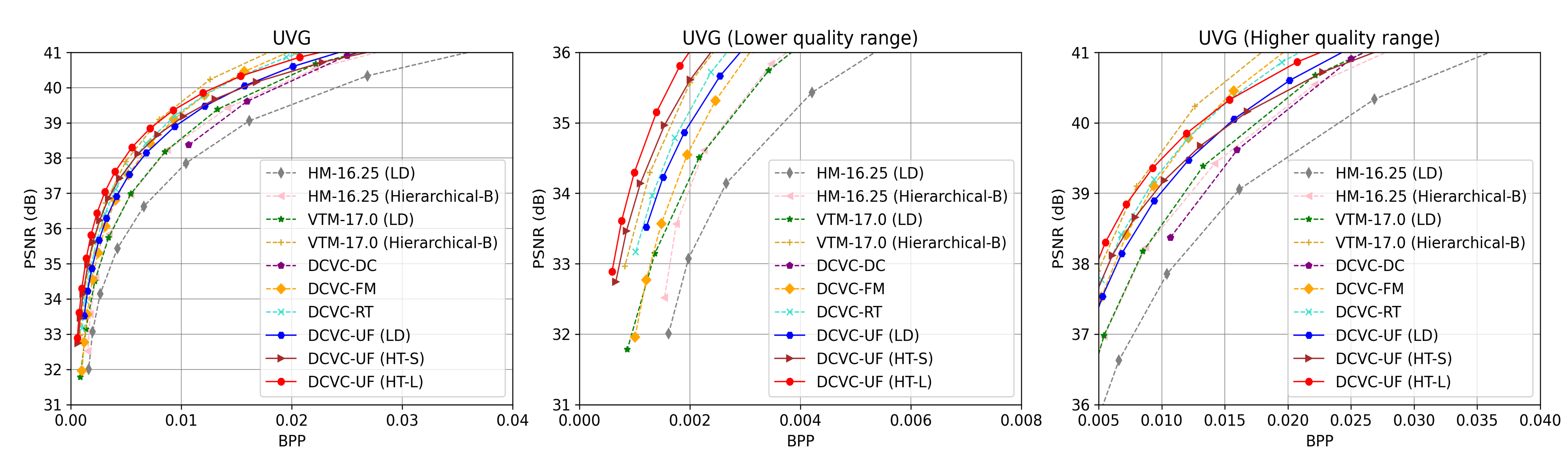}
    \vspace{-6mm}
    \caption{Rate-distortion curves on UVG dataset. BPP means bits per pixel.  More curves are in the supplementary material.
    }
    \vspace{-3mm}
  \label{fig_yuv_psnr_allf_curve}
\end{figure*}
\section{Experimental Results}

\subsection{Experimental Settings}
\textbf{Implementation Details}. For delay-relaxed scenario, our high-throughput (HT) codec DCVC-UF uses a chunk size of $N=8$. We provide two network scales—DCVC-UF (HT-S) and DCVC-UF (HT-L)—denoting small and large models, respectively. To enable low-delay (LD) operation, the framework also supports  $N=1$ (single-frame chunk), i.e., DCVC-UF (LD).\\
\textbf{Training Details}. Following \cite{DCVC-FM}, we first train DCVC-UF codecs on the ready-made 7-frame Vimeo-90k \cite{vimeo} dataset, then fine-tune using longer sequences generated from original Vimeo videos \cite{ori_vimeo}. As discussed in Section \ref{sec_long}, our chunk coding enables training on $512\times512$  videos with up to 1024 frames. However, assembling diverse, high-quality videos of such length is still challenging. Therefore, we currently fine-tune with 128-frame sequences and leave the exploration of longer training datasets for future work.\\ 
\textbf{Testing Details}. We evaluate on HEVC Class B$\sim$E \cite{flynn16common}, UVG \cite{uvg}, and MCL-JCV \cite{mcl-jcv}. For traditional codecs, we compare with HM \cite{HM} and VTM \cite{VTM}, representing the best H.265 and H.266 encoders, respectively. Configuration details are in the supplementary material. For NVCs, we compare with previous SOTA DCVC series \cite{DCVC-DC,DCVC-FM,DCVC-RT}, with all models tested using actual bit-stream writing and decoding. Compression ratio is measured by BD-Rate \cite{bd-rate}, where positive values indicate a bitrate increase and negative values indicate savings. Video quality is reported using PSNR, with all frames evaluated in YUV420 colorspace. For both low-delay and delay-relaxed settings, the intra-period is set to --1 for all codecs to present their best compression ratios. We measure coding speed on a single GPU, using a sequential chunk-by-chunk coding process (chunk size $N=8$ for HT and $N=1$ for LD). We currently do not employ cross-chunk pipeline parallelism (e.g., overlapping the network inference and entropy coding of different chunks), indicating potential for further acceleration.

\subsection{Comparisons with Previous SOTA Methods}

Table \ref{tab:compare_yuv_allf} shows the BD-Rate comparison. In terms of the averaged bitrate saving, our DCVC-UF (HT-L) achieves the best performance, i.e., an average of 42.2\% bitrate saving over VTM (LD). By contrast, the corresponding result of the VTM (Hierarchical-B) is 33.8\%, where we use the default GOP size 32 and its maximum frame delay is 31 frames. Our chunk size is 8, leading to a maximum 7 frame delay, which is much smaller. If VTM (Hierarchical-B) also uses GOP size 8, the corresponding bitrate saving is reduced to 23.7\%.  It shows the compression efficiency of our chunk coding.

 \begin{table}[t]
  \centering
  \caption{Ablation study.  VTM-17.0 (LD) is as the anchor for BD-Rate results. Decoding FPS is tested on 4090 GPU.}
   \renewcommand{\arraystretch}{1.8}
    \small
    \resizebox{1.03\linewidth}{!}{
      \begin{tabular}{ l | l c c }
            \toprule
            ID & Settings & BD-Rate & \makecell[c]{Decoding\\FPS}\\
            \hline
            A & DCVC-RT $\to$ \textit{Anchor}                                   & -21.0\% & 105.3\\ [0.5ex] 
            B & \makecell[l]{A + Chunk coding \\(w/o frame-specific decoder)}   & -10.1\% & 349.1\\ [2ex] 
            C & \makecell[l]{A + Chunk coding \\(w/ frame-specific decoder)}    & -25.3\% & 343.2\\
            D & C + Streamlined entropy model                                   & -23.4\% & 453.3\\
            E & D + Training with 128-frame video                               & -31.6\% & 453.3\\ [-1.6ex]
              & $\to$ \textit{Proposed DCVC-UF (HT-S)}    &   &  \\
            \bottomrule
        \end{tabular}
    }
    \vspace{-0.3cm}
  \label{aba_table}
\end{table}

\begin{table*}[t]
    \caption{Complexity analysis. The encoding/decoding speeds (frames per second, FPS) are evaluated across various resolutions and devices. Average BD-Rate results are presented using VTM-17.0 (LD) as the anchor. MACs are tested on 1080p videos. OOM means out-of-memory.}
    \vspace{-1mm}
    \centering
    \setlength{\tabcolsep}{6pt}

    \begin{tabular}{cc}
        \begin{minipage}{0.43\textwidth} 
          \renewcommand{\arraystretch}{1.8}
            \centering
            \resizebox{0.92\linewidth}{!}{
        	\begin{tabular}{ l | c | c | c }
                \toprule
        		Model & Average BD-Rate &Average MACs/frame & Params\\
        	    \midrule
                DCVC-FM    & $-21.3\%$ & 2642G & 18.3M \\
        	    \midrule
                DCVC-RT   & $-21.0\%$ & 385G & 20.7M \\
        	    \midrule
                DCVC-UF (LD)   & $-9.5\%$ & 170G & 9.7M \\
                \midrule
                DCVC-UF (HT-S)   & $-31.6\%$ & 211G & 81.2M \\
                \midrule
                DCVC-UF (HT-L)   & $-42.2\%$ & 343G & 120.5M \\
                \bottomrule
        	\end{tabular}
            }
            \vspace{10px}
            \subcaption{Computational complexity and BD-Rate.} 
            \vspace{3px}
        \end{minipage}
        & 
        \hspace{-15px}
        \begin{minipage}{0.5\textwidth}
          \renewcommand{\arraystretch}{1.9}
            \centering
            \resizebox{\linewidth}{!}{
              \begin{tabular}{ l | c  | c | c | c | c}
                \toprule
        		Model & 2080Ti & 4090 & A100 & H100 & B200\\
                \midrule
                DCVC-FM  & 4.0 / 4.7 & 9.3 / 10.4  & 8.5 / 9.4 & 11.2/16.9 & 12.3/21.7 \\
        	    \midrule
                DCVC-RT   &73.3 / 67.0 &225.1 / 185.2& 173.9 / 149.2 & 284.4/252.4 & 289.7/263.9  \\
        	    \midrule
                DCVC-UF (LD) & 191.0/203.7 & 432.5/634.6 & 525/501.2 & 777.2/706.5 & 848.4/902.5 \\
                \midrule
                DCVC-UF (HT-S) & 348.1/251.8 & 1194.5/985.2 & 1098.4/786.1 & 1901.5/1347.2 & 2318.2/1633.1\\
                \midrule
                DCVC-UF (HT-L) & 206.4/163.1 & 778.9/558.4 & 648.2/459.4 & 1083.1/752.0 &  1424.9/908.3\\
                \bottomrule
        	\end{tabular}
    }
            \vspace{3px}
            \subcaption{Coding speed on $1280\times720$ videos.} 
            \vspace{3px}
        \end{minipage} \\
        
        \hspace{-14px}
        \begin{minipage}{0.5\textwidth}
          \renewcommand{\arraystretch}{1.93}
            \centering
            \resizebox{\linewidth}{!}{
              \begin{tabular}{ l | c  | c | c | c | c}
                \toprule
        		Model & 2080Ti & 4090 & A100 & H100 & B200\\
                \midrule
                DCVC-FM  & 1.9/2.3 & 3.7/4.4  & 5.0/5.9 & 7.4/8.3 & 8.2/11.9 \\
        	    \midrule
                DCVC-RT   &39.5/34.1 &118.8/105.3& 125.2/112.8 & 214.9/184.7 & 220.3/211.7  \\
        	    \midrule
                DCVC-UF (LD) & 103.0/105.4 & 313.6/353.8 & 317.0/314.5 & 503.8/505.1 & 625.6/621.9 \\
                \midrule
                DCVC-UF (HT-S) & 171.6/123.2 & 655.9/453.3 & 576.2/411.1 & 957.4/668.9 & 1415.1/945.8 \\
                \midrule
                DCVC-UF (HT-L) & 109.0/87.4 & 371.1/273.6 & 331.0/247.4 & 512.4/394.1 & 811.7/551.6  \\
                \bottomrule
        	\end{tabular}
            }
            \vspace{2px}
            \subcaption{Coding speed on $1920\times1080$ videos.} 
        \end{minipage}
        & 
        \hspace{-15px}
        \begin{minipage}{0.5\textwidth}
        \renewcommand{\arraystretch}{1.82}
            \centering
            \resizebox{\linewidth}{!}{
        	\begin{tabular}{ l | c  | c | c | c | c}
                \toprule
        		Model & 2080Ti & 4090 & A100 & H100 & B200\\
                \midrule
                DCVC-FM  & OOM & OOM  & 1.0/1.2 & 1.8/2.2 & 2.5/3.3 \\
        	    \midrule
                DCVC-RT   &11.6/9.9 &29.9/26.5& 35.5/29.5 & 56.9/52.0 & 91.6/87.1  \\
        	    \midrule
                DCVC-UF (LD) & 29.2/29.9 & 80.2/81.8 & 91.6/93.6 & 156.8/158.2 & 230.2/226.4 \\
                \midrule
                DCVC-UF (HT-S) & 45.9/31.6 & 139.5/94.7 & 155.3/107.3 & 255.9/179.6 & 424.0/289.5\\
                \midrule
                DCVC-UF (HT-L) & 26.1/21.9 & 83/61.9 & 84.3/67.4 & 129.0/99.8 & 237.9/177.2  \\
                \bottomrule
        	\end{tabular}
        }
            \vspace{3px}
            \subcaption{Coding speed on $3840\times2160$ videos.} 
        \end{minipage}
    \end{tabular}
  \label{tab:compare_complexity}
  \vspace{-3mm}
\end{table*}

In terms of the actual encoding and decoding speeds on 4090 GPU, DCVC-UF (HT-L) can achieve 371.1 FPS and 273.6 FPS for 1080p videos, respectively. When using a smaller network structure, our DCVC-UF (HT-L) can boost the  encoding and decoding speeds to 655.9 FPS and 453.3 FPS, respectively. But the average bitrate saving can still keep 31.6\%, comparable with VTM (Hierarchical-B) using GOP 32. This shows the advanced rate-distortion-complexity trade-off of our chunk coding paradigm. For the low-delay setting, our DCVC-UF (LD) can reach 313.6 encoding and 353.8 decoding speeds. Although the bitrate saving over VTM (LD) is lower than that of DCVC-RT, the decoding speed acceleration is more than $3\times$ times. 

Figure \ref{fig_yuv_psnr_allf_curve} illustrates the rate-distortion curves on the UVG dataset. In the lower quality range, both DCVC-UF (HT-L) and DCVC-UF (HT-S) consistently outperform all previous codecs. At a higher quality range, VTM (Hierarchical-B) shows better performance, which is expected given our models' lightweight nature. However, this performance gap primarily manifests above 40 dB, where quality differences become imperceptible to human vision.

\subsection{Ablation Study}

Table \ref{aba_table} presents an ablation study. Starting from DCVC-RT as our baseline (A), we first introduce chunk coding without frame-specific decoders (B), which significantly boosts decoding speed from 105.3 to 349.1 FPS but compromises compression efficiency (bitrate saving changes from 21.0\% to 10.1\%). This drop highlights the challenge of using a single unified decoder for all temporal positions. Adding our frame-specific decoders (C) improves compression performance to 25.3\% bitrate saving while maintaining high decoding speed at 343.2 FPS, validating our design of specialized decoders for each temporal position. With the streamlined entropy model (D), decoding reaches 453.3 FPS at 23.4\% bitrate saving, evidencing the acceleration benefit of the single-step bit-stream interaction. Finally, extending training to 128-frame sequences (E) substantially improves bitrate saving to 31.6\% without affecting decoding speed, confirming that our chunk-based framework enables efficient learning of long-term temporal correlations. 

\subsection{Throughput Scaling on General-Purpose GPUs}

Traditional video codecs are tightly coupled to hardware, often requiring multi-year standardization cycles and bespoke silicon or platform-specific hand-tuning. In contrast, neural codecs rely on commodity GPU primitives (e.g., convolutions and matrix multiplications) that naturally benefit from progress in AI accelerators. This compute alignment eliminates most hardware-specific engineering across heterogeneous devices, enabling automatic speedups as either model architectures or GPU hardware improves.
Table~\ref{tab:compare_complexity} demonstrates NVC’s scalability across GPU generations. On the B200, DCVC-UF (HT-S) reaches 1415.1 encoding and 945.8 decoding FPS for 1080p videos, setting a new NVC speed record in history. These gains show the potential of NVCs for large-scale enterprise workloads (e.g., cloud video analytics, mass transcoding). The consistent improvements from consumer GPUs (e.g., 2080 Ti, 4090) to datacenter accelerators (e.g., H100, B200) validate its ability to harness general-purpose compute advances without GPU-specific re-engineering.

\section{Conclusion and Limitation}

This paper presents DCVC-UF NVC that achieves unprecedented encoding and decoding speeds while maintaining high compression efficiency. DCVC-UF adopts a chunk-based coding framework that processes multiple frames simultaneously. It not only dramatically enhances coding throughput but also facilitates more effective modeling of long-term temporal correlations.
By introducing frame-specific decoders that act as specialized experts for each temporal position, our approach better adapts to diverse video content characteristics. Furthermore, our streamlined entropy model consolidates bit-stream interactions into a single step by decoupling scale and mean estimation, markedly reducing operational overheads. DCVC-UF represents a significant step in the evolution of NVC, fundamentally transforming the commonly-used yet complex hierarchical-B coding to a much more efficient chunk coding.

Despite these advances, the current DCVC-UF uses a fixed chunk size, which may not be optimal for videos with varying temporal characteristics. Future work could explore adaptive chunk sizing based on content complexity.
{
    \small
    \bibliographystyle{ieeenat_fullname}
    \bibliography{main}
}

\clearpage
\maketitlesupplementary
\appendix
This document provides supplementary material for our paper, detailing the experimental settings and additional results for our proposed ultra-fast neural video codec (NVC), DCVC-UF (Ultra-Fast).

\section{Test Settings}
This section details the experimental configurations used for comparing our NVC with traditional codecs across both YUV420 and RGB (shown in Section \ref{rgb_results}) colorspaces.

\textbf{YUV420 Colorspace.} The YUV420 colorspace is the standard for most practical video applications and traditional codecs, which are highly optimized for this format. Therefore, benchmarking in YUV420 is crucial for assessing the practical performance of NVCs against established standards.
For traditional codecs, we use the reference software for H.265 (HM \cite{HM}) and H.266 (VTM \cite{VTM}). In the low-delay setting, we use the \textit{encoder\_lowdelay\_main10.cfg} and \textit{encoder\_lowdelay\_vtm.cfg} configurations for HM and VTM, respectively. For the delay-relaxed setting, we use \textit{encoder\_randomaccess\_main10.cfg} and \textit{encoder\_randomaccess\_vtm.cfg}. The parameters for each video are as follows:
\begin{itemize}
	\item 
	-c \{{\em config file name}\}\par
	-\/-InputFile=\{{\em input video name}\}\par
	-\/-InputBitDepth=8\par
	-\/-OutputBitDepth=8 \par
	-\/-OutputBitDepthC=8 \par
	-\/-FrameRate=\{{\em frame rate}\}\par
	-\/-DecodingRefreshType=2\par
	-\/-FramesToBeEncoded=\{{\em frame number}\}\par
	-\/-SourceWidth=\{{\em width}\}\par
	-\/-SourceHeight=\{{\em height}\}\par
	-\/-IntraPeriod=\{{\em intra period}\}\par
	-\/-QP=\{{\em qp}\}\par
	-\/-Level=6.2\par
	-\/-BitstreamFile=\{{\em bitstream file name}\}\par
\end{itemize}

For both low-delay and delay-relaxed settings, it should be noted that we set the intra-period to --1 for HM and VTM to report their best compression ratio. Here we show a comparison for VTM under the delay-relaxed setting: if using VTM with intra-period=32 as the anchor, VTM with intra-period=--1 can achieve an average of  19.8\% bitrate saving on the six test datasets (HEVC Class B$\sim$E, UVG, and MCL-JCV). 

\textbf{RGB Colorspace.} Since our test datasets are originally in the YUV420 format, we convert them to RGB for this evaluation. Following the methodology of JPEG AI \cite{jpeg_ai,jpeg_ai2} and \cite{DCVC-DC}, we use the BT.709 standard for the YUV-to-RGB conversion. This is because
using BT.709 obtains higher compression ratio under the similar visual quality when compared with the BT.601. As demonstrated in \cite{DCVC-DC}, traditional codecs achieve superior compression when encoding RGB content by using an internal 10-bit YUV444 pipeline,  even though the final distortion is measured in RGB. We adopt this same best-practice configuration for our tests.
For HM and VTM, we use the \textit{encoder\_lowdelay\_main\_rext.cfg} and \textit{encoder\_lowdelay\_vtm.cfg} configurations for the low-delay scenario. 
The parameters for each video are as follows:
\begin{itemize}
	\item 
	-c \{{\em config file name}\}\par
	-\/-InputFile=\{{\em input file name}\}\par
	-\/-InputBitDepth=10\par
	-\/-OutputBitDepth=10 \par
	-\/-OutputBitDepthC=10 \par
	-\/-InputChromaFormat=444\par
	-\/-FrameRate=\{{\em frame rate}\}\par
	-\/-DecodingRefreshType=2\par
	-\/-FramesToBeEncoded=\{{\em frame number}\}\par
	-\/-SourceWidth=\{{\em width}\}\par
	-\/-SourceHeight=\{{\em height}\}\par
	-\/-IntraPeriod=\{{\em intra period}\}\par
	-\/-QP=\{{\em qp}\}\par
	-\/-Level=6.2\par
	-\/-BitstreamFile=\{{\em bitstream file name}\}\par
\end{itemize}

For both YUV420 and RGB evaluations, all traditional codecs were configured with their best settings and reference structures to ensure rigorous comparison.

\begin{table*}[t]
    \caption{BD-Rate (\%) comparison in RGB colorspace. All frames are tested. }
    \vspace{-0mm}
    \centering
    \renewcommand\arraystretch{1.8}
    \setlength{\tabcolsep}{3.5pt}
    \resizebox{0.9\linewidth}{!}{
    \begin{threeparttable}
    	\begin{tabular}{ l c c c c c c c c c }
            \toprule
    		\multirow{2}{*}{Method} & \multirow{2}{*}{UVG} & \multirow{2}{*}{MCL-JCV} & \multirow{2}{*}{HEVC B} & \multirow{2}{*}{HEVC C} & \multirow{2}{*}{HEVC D} & \multirow{2}{*}{HEVC E} & \multirow{2}{*}{Average} & \multicolumn{2}{c}{Coding Speed} \\
            ~ & ~ & ~ & ~ & ~ & ~ & ~ & ~ & Enc. & Dec. \\
            \hline    
            \textit{Low-Delay (LD) Codecs}\\
            \hline
            VTM-17.0 (LD)                       & $0.0$   & $0.0$   & $0.0$   & $0.0$   & $0.0$   & $0.0$   & $0.0$  & $0.01$ FPS  & $23.6$ FPS\\
            \hline
            HM-16.25 (LD)                       & $43.2$   & $49.5$   & $49.9$   & $45.2$   & $39.9$   & $47.7$   & $45.9$  & $0.05$ fps & 39.6 fps\\
            \hline
            DCVC-DC                             & $9.2$   & $0.0$   & $14.9$   & $5.3$   & $-7.8$   & $87.7$   & $18.2$ & $2.3$ FPS  & $2.9$ FPS \\
            \hline    
            DCVC-FM                             & $-10.4$   & $-1.1$   & $-11.2$   & $-26.5$   & $-33.7$   & $-12.1$   & $-15.8$  & $3.7$ FPS  & $4.4$ FPS  \\
            \hline
            DCVC-RT                             & $-17.2$   & $-6.8$   & $-11.3$   & $-15.8$   & $-21.3$   & $-11.4$   & $-14.0$  & $118.8$ FPS& $105.3$ FPS \\
            \hline    
            \textbf{DCVC-UF (LD)}               & $-7.2$ & $8.6$  & $2.2$ & $-1.5$ & $-10.3$ & $-4.0$   & $-2.0$ & $\mathbf{313.6}$ FPS & $\mathbf{353.8}$ FPS \\
            \hline    
            \textit{Delay-Relaxed Codecs}\\
            \hline
            \textbf{DCVC-UF (HT-S)}      & $-22.7$ & $-3.1$ & $-13.0$ & $-24.8$ & $-37.3$ & $-52.7$ & $-25.6$ & $\mathbf{655.9}$ FPS & $\mathbf{453.3}$ FPS \\        
            \hline
           \textbf{DCVC-UF (HT-L)}       & $-34.5$ & $-16.2$ & $-29.4$ & $-37.4$ & $-47.5$ & $-57.7$ & $-37.1$ & $\mathbf{371.1}$ FPS & $\mathbf{273.6}$ FPS \\
            \bottomrule
    	\end{tabular}
        \begin{tablenotes}
            \item  Intra-period=--1 for all codecs and settings. The coding speeds of NVCs are tested on $1920 \times 1080$ videos with 4090 GPU. 
        \end{tablenotes}
    \end{threeparttable}
	}
  \label{tab:compare_rgb_allf}
  \vspace{-0mm}
\end{table*}

\section{Depth-wise Convolution Block Details}
\begin{figure}[t]
  \centering
    \includegraphics[width=\linewidth]{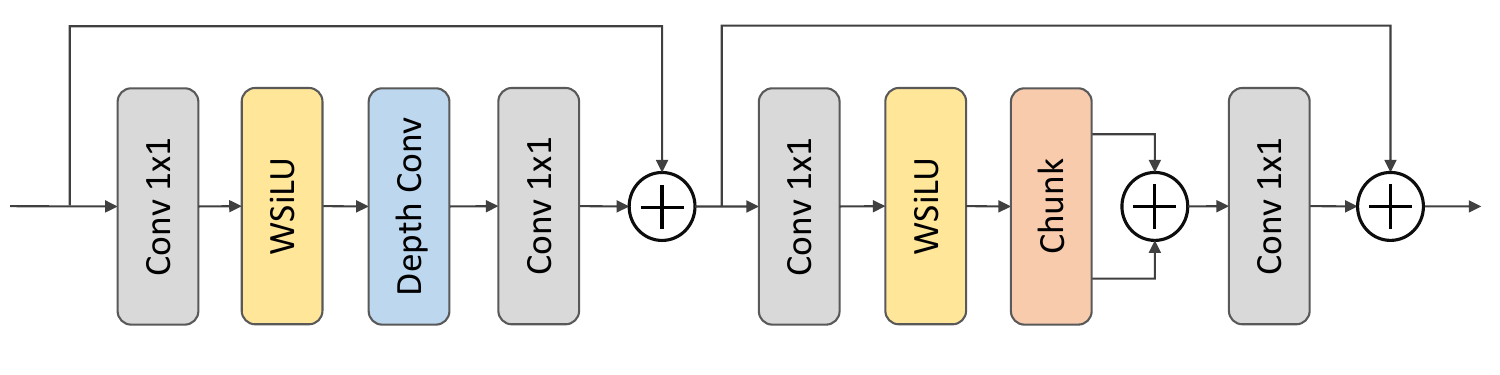}
    \caption{ The structure of DC Block (depth-wise convolution block) \cite{DCVC-RT}.
    }
  \label{fig_dc_framwork}
\end{figure}

Our DCVC-UF architecture is deliberately built from a single, lightweight primitive: depth-wise convolution (DC) blocks (as illustrated in Fig. \ref{fig_dc_framwork}) \cite{DCVC-RT}, which serve as the fundamental building units across all modules. By avoiding complex operations or heterogeneous specialized blocks, the overall network remains conceptually simple and highly efficient to implement. For the delay-relaxed setting with chunk size $N=8$, our high-throughput model DCVC-UF (HT-S) employs 6, 7, and 11 DC blocks in the chunk encoder, chunk decoder, and cross-chunk context generation modules, respectively. Each of the 8 frame-specific decoders (corresponding to the 8 frames in the chunk) contains 3 DC blocks, enabling all frames to be reconstructed in parallel. The larger configuration DCVC-UF (HT-L) increases the numbers of DC blocks to 7, 11, and 12 for the chunk encoder, chunk decoder, and cross-chunk context generation modules, respectively, and allocates 5 DC blocks to each of the 8 frame-specific decoders, providing stronger modeling capacity at modest additional complexity. For the low-delay setting with chunk size $N=1$, DCVC-UF (LD) adopts a more compact design with 3, 3, and 9 DC blocks in the chunk encoder, chunk decoder, and cross-chunk context generation modules, respectively, and a single frame decoder comprising 3 DC blocks, yielding an efficient instantiation tailored to low-delay applications.

\section{Results on RGB colorspace}
\label{rgb_results}
Our DCVC-UF framework also employs a unified YUV444 colorspace that serves as a universal interface for both input and output. For both YUV420 and RGB videos, they will be converted into YUV444 for inference and converted back for loss calculation. The unified approach enhances practical deployment by allowing a single trained model to efficiently process multiple video formats without requiring specialized model variants for different formats, thereby reducing storage requirements and deployment complexity.  The model is also  trained exclusively on YUV444 colorspace, which greatly simplifies the training process.  Table \ref{tab:compare_rgb_allf} presents the performance comparisons in the RGB colorspace. From this table, we can find that our DCVC-UF models achieve the advanced rate-distortion-complexity trade-off. These results validate the effectiveness of DCVC-UF for practical video compression applications.

\begin{figure*}[p]
\vspace{-4mm}
  \centering
    \includegraphics[width=\linewidth]{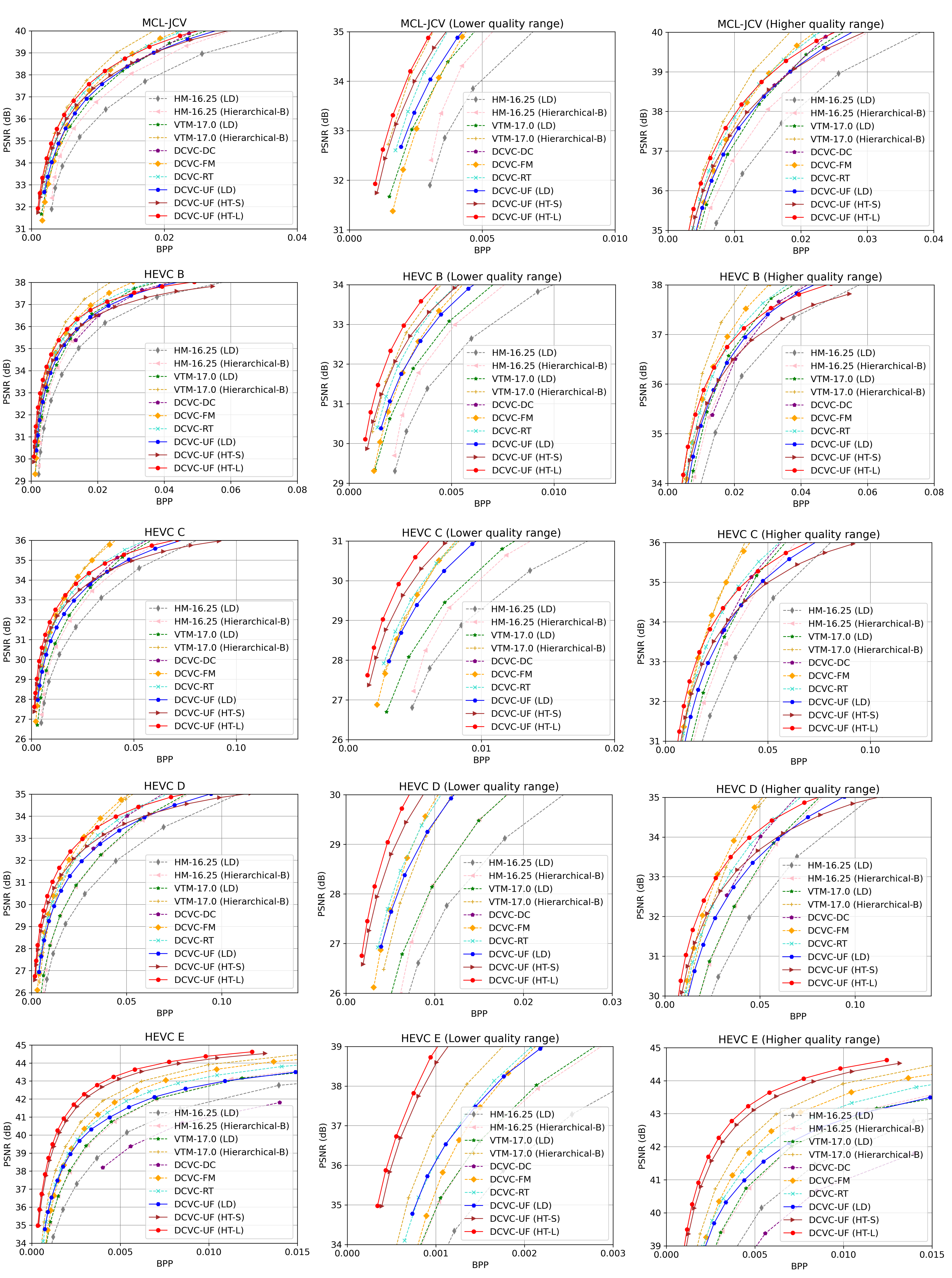}
    \vspace{-8.5mm}
    \caption{Rate-distortion curves.
    }
    \vspace{-3mm}
  \label{supp_fig_yuv_psnr_allf_curve}
\end{figure*}

\section{Rate-Distortion Curves}
This section presents comprehensive rate-distortion (RD) curves across multiple datasets in the YUV420 colorspace, with all frames tested under the intra-period=--1 configuration. As illustrated in Fig. \ref{supp_fig_yuv_psnr_allf_curve}, our analysis reveals different performance characteristics across different quality ranges and datasets. In the lower quality range, our proposed DCVC-UF (HT-L) demonstrates the best compression efficiency on all datasets, consistently outperforming both traditional codecs (H.265/HM, H.266/VTM) and previous state-of-the-art neural video codecs. This advantage is particularly pronounced in scenarios where bandwidth constraints or storage costs are critical, making our approach highly suitable for real-world applications.

In the higher quality range, the performance landscape becomes more heterogeneous. Our DCVC-UF (HT-L) achieves the best compression ratio on the HEVC E dataset, demonstrating its capability to handle specific content types effectively. However, on other datasets, VTM with Hierarchical-B configuration maintains its advantage in the very high quality range. This performance gap is expected, as it represents a deliberate design trade-off: our models prioritize computational efficiency while maintaining high compression ratio. The lightweight architecture that enables ultra-fast encoding and decoding speeds inherently limits the model capacity available for achieving maximum compression ratio at very high quality levels. These results underscore the practical applicability of our approach for scenarios requiring very fast processing with good compression ratio, while also identifying opportunities for future improvements in the high-quality domain through increased model capacity or more advanced design.

\end{document}